\pdfoutput=1
\documentclass[11pt]{article}

\usepackage{EMNLP2022}

\usepackage{times}
\usepackage{latexsym}

\makeatletter
\newcommand{\printfnsymbol}[1]{%
  \textsuperscript{\@fnsymbol{#1}}%
}
\makeatother

\usepackage[T1]{fontenc}
\usepackage[utf8]{inputenc}

\usepackage{microtype}

\usepackage{inconsolata}

\usepackage{amsmath}
\usepackage{amssymb}
\usepackage[capitalise]{cleveref}
\usepackage{blindtext}
\usepackage{graphicx}
\usepackage{multirow}
\usepackage{booktabs}
\usepackage{subfigure}
\usepackage{stfloats}

\usepackage{paralist}
\usepackage[shortlabels]{enumitem}

\usepackage{xspace}
\newcommand*{\eg}{e.g.\@\xspace}
\newcommand*{\ie}{i.e.\@\xspace}

\newcommand{\ModelName}{EMAT\xspace}
\newcommand{\ModelNames}{EMATs\xspace}

\newcommand{\exper}[1]{\textsc{#1}}
\newcommand{\prefix}{\exper{Prefix}\xspace}

\usepackage{glossaries}
\glsdisablehyper
\newacronym{ODQA}{ODQA}{Open-Domain Question Answering}

\newif\ifcomments
\newcounter{psCounter}
\newif\ifpsvar
\ifcomments
    \psvartrue
\fi
\ifpsvar
    \newcommand{\ps}[1]{{\small \color{red} \refstepcounter{psCounter}\textsf{[PS]$_{\arabic{psCounter}}$:{#1}}}}
\else
    \newcommand{\ps}[1]{}
\fi

\newcounter{ywCounter}
\newif\ifywvar
\ifcomments
    \ywvartrue
\fi
\ifpsvar
    \newcommand{\yw}[1]{{\small \color{cyan} \refstepcounter{ywCounter}\textsf{[YW]$_{\arabic{ywCounter}}$:{#1}}}}
\else
    \newcommand{\yw}[1]{}
\fi

\newcounter{pmmCounter}
\newif\ifywvar
\ifcomments
    \ywvartrue
\fi
\ifpsvar
    \newcommand{\pmm}[1]{{\small \color{red} \refstepcounter{pmmCounter}\textsf{[PM]$_{\arabic{pmmCounter}}$:{#1}}}}
\else
    \newcommand{\pmm}[1]{}
\fi

\title{An Efficient Memory-Augmented Transformer for \\Knowledge-Intensive NLP Tasks}

\author{
  Yuxiang Wu \printfnsymbol{2} \qquad Yu Zhao \printfnsymbol{3} \qquad Baotian Hu \printfnsymbol{3}$^{*}$ %\thanks{ \enspace   Corresponding author.} 
  \qquad Pasquale Minervini \printfnsymbol{4}\printfnsymbol{2} \\
  \bf Pontus Stenetorp \printfnsymbol{2} \qquad Sebastian Riedel \printfnsymbol{2} \\
  \printfnsymbol{2} University College London, London, UK \quad 
    \printfnsymbol{3} Harbin Institute of Technology, Shenzhen, PRC \\
  \printfnsymbol{4} University of Edinburgh, Edinburgh, UK \\ 
  { \normalsize \tt \{yuxiang.wu, p.stenetorp, s.riedel\}@cs.ucl.ac.uk } \qquad { \normalsize \tt  p.minervini@ed.ac.uk} \\
  { \normalsize \tt 20s151163@stu.hit.edu.cn \qquad hubaotian@hit.edu.cn}
}

\begin{document}
\maketitle
\begingroup\def\thefootnote{*}\footnotetext{Corresponding author.}\endgroup

\begin{abstract}
Access to external knowledge is essential for many natural language processing tasks, such as question answering and dialogue.
Existing methods often rely on a \emph{parametric model} that stores knowledge in its parameters, or use a \emph{retrieval-augmented model} that has access to an external knowledge source.
Parametric and retrieval-augmented models have complementary strengths in terms of computational efficiency and predictive accuracy.
To combine the strength of both approaches, we propose the Efficient Memory-Augmented Transformer (\ModelName) -- it encodes external knowledge into a key-value memory and exploits the fast maximum inner product search for memory querying.
We also introduce pre-training tasks that allow \ModelName to encode informative key-value representations, and to learn an implicit strategy to integrate multiple memory slots into the transformer.
Experiments on various knowledge-intensive tasks such as question answering and dialogue datasets show that, simply augmenting parametric models (T5-base) using our method produces more accurate results (\eg, $25.8 \rightarrow 44.3$ EM on NQ) while retaining a high throughput (\eg, 1000 queries/s on NQ).
Compared to retrieval-augmented models, \ModelName runs substantially faster across the board and produces more accurate results on WoW and ELI5.\footnote{Our code and datasets are available at \url{https://github.com/uclnlp/EMAT}.}
\end{abstract}

\section{Introduction}

% \begin{figure}[htb]
% \begin{center}
% %\includegraphics[trim={2cm 5cm 4cm 5cm},clip,width=0.9\columnwidth]{figures/splash-crop.pdf}
% \includegraphics[width=0.9\columnwidth]{figures/splash-crop.pdf}
% \end{center}
% \caption{\yw{Placeholder for a splash figure}}
% \label{fig:splash}
% \end{figure}

NLP tasks often require knowledge that is not explicitly provided with the input.
For example, Open-Domain Question Answering~(ODQA) requires answering an open-domain question without given context passages~\citep{DBLP:conf/acl/ChenFWB17}, and likewise for open-domain dialogue~\citep{WoW}.
To handle such tasks, one key challenge is storing and accessing potentially large amounts of knowledge. %that the NLG models have to memorise a large amount of knowledge.
% 
% Parametric
One approach is a parametric method that trains a sequence-to-sequence generator to represent knowledge within model parameters.
% \emph{Parametric models} use knowledge stored in their parameters to solve a task: this is computationally efficient at inference time, but does not provenance or transparency into the process, and models can suffer from significant knowledge gaps~\citep{LAMA,DBLP:conf/eacl/LewisSR21,DBLP:journals/corr/abs-2109-01156}.
%
%This method is fast at inference time, but it limited in terms of knowledge coverage efficient~\citep{LAMA}.
%
\citet{LAMA} find that pre-trained Language Models~(PLMs) learn a partial knowledge base in their parameters, but its coverage is limited.
Increasing model size can improve this issue~\citep{T5,DBLP:conf/emnlp/RobertsRS20,GPT3}; % however, larger language models require significantly more computation, and hence this method is not scalable to large-scale knowledge.
however, larger language models require significant computational resources. %making it challenging for most academics, students, and researchers to use them
% which can introduce challenges in terms of accessibility and environmental impact~\citep{DBLP:journals/cacm/SchwartzDSE20}.
%
% Recent works with generative PLMs are more capable in memorising factual knowledge~\citep{T5, GPT3, Gopher, PaLM}, and perform better on knowledge intensive natural language tasks such as question answering and dialogue system.
%

% Retrieval-augmented generator
% We went with parametric first, now let's go with Semi-parametric
%
%Another approach is retrieval-augmented generator~\cite{RAG,FiD} (a.k.a., semi-parametric model), which retrieves relevant passages from an external knowledge source (\eg, Wikipedia), and then feeds the retrieved text as part of the input to the encoder to inform generation. 
%
\emph{Retrieval-augmented models}~\citep{REALM, RAG, FiD, spanlp-2022-semiparametric}, on the other hand, 
% allow the number of model parameters to grow based on the quantity of available data~\citep{spanlp-2022-semiparametric}: such models can, for example, 
retrieve relevant passages from an external knowledge source~(\eg, Wikipedia), and use the retrieved passages to inform generation.
Despite being more accurate, retrieval-augmented models are often significantly more costly computation-wise than their parametric counterparts, since they require retrieving, encoding, and integrating the external knowledge at inference time.
%

\begin{figure*}[t]
\begin{center}
\includegraphics[width=0.9\textwidth]{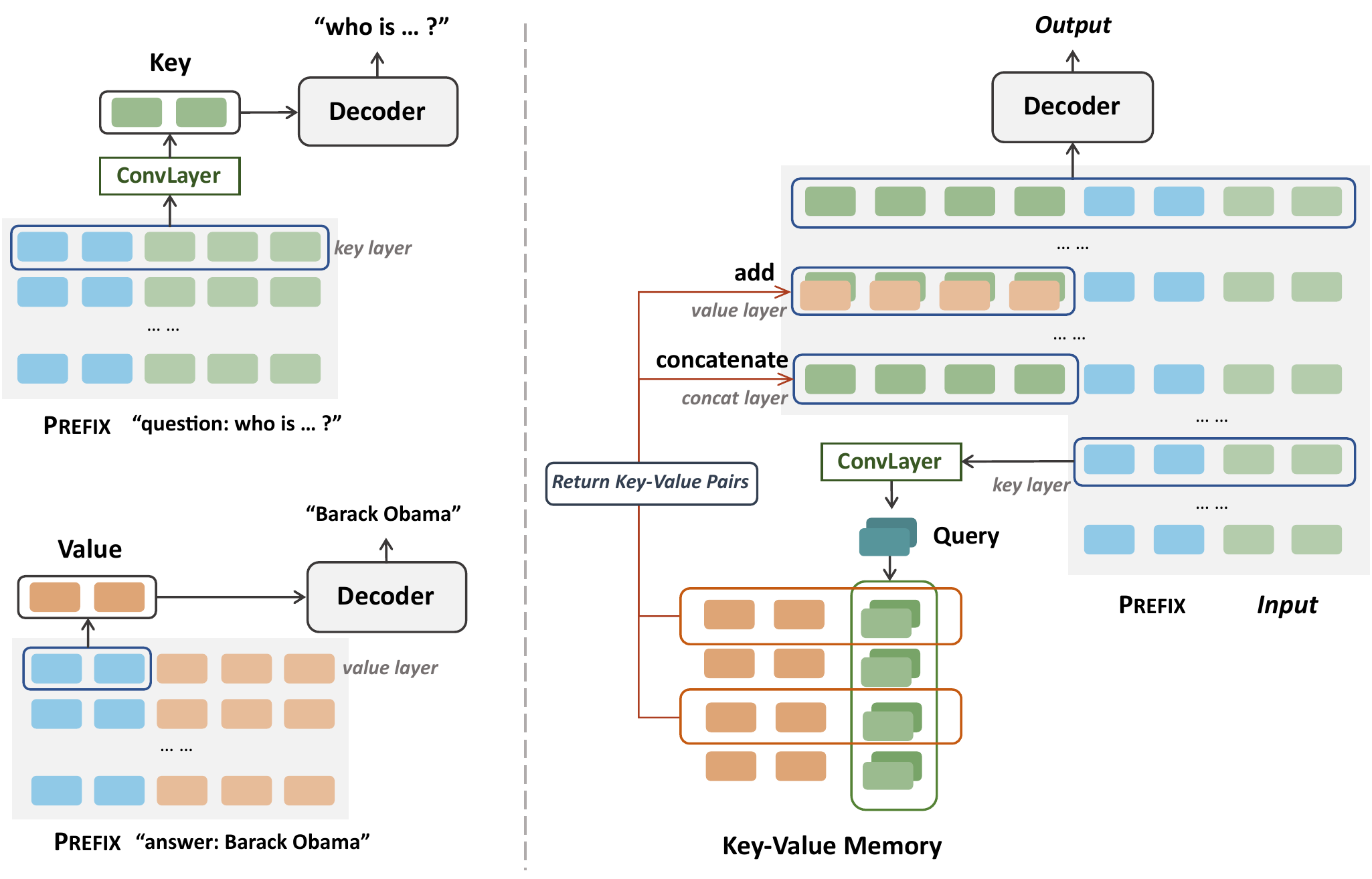}
\end{center}
\caption{
%The architecture of our proposed \ModelName. \pmm{Too concise! expand}
Architecture of the proposed Efficient Key-Value Memory Augmented Transformers (EMAT): factual knowledge is stored in a key-value memory (\cref{sec:kvm}) where keys and values correspond to questions and answers, respectively;
during inference, the model retrieves information from the memory via MIPS (\cref{sec:retriever}) and uses it to condition the generation process.
}
\label{fig:arch}
\end{figure*}

%
%To get the best of both worlds, 
% we investigate methods that efficiently scale-up parametric models, and allow text-to-text generators to access knowledge without substantial delay.
%
To combine the strengths of both parametric and retrieval-augmented models, we propose Efficient Memory-Augmented Transformers~(\ModelNames) --
an extension to Transformer-based models augmented with an efficient key-value memory module.
% Overview of our method
\ModelName first encodes the external knowledge source into key embeddings and value embeddings, to construct the key-value memory~(\cref{sec:kvm}).
We choose PAQ~\citep{paq}, a large collection of question-answering generated from Wikipedia, as our knowledge source; and we encode the questions as keys and answers as values.
The transformer model produces dense query vector, retrieves from the key-value memory~(\cref{sec:retriever}), and integrates the returned dense key-value vectors at different encoder layers to enhance generation~(\cref{sec:integration}). 
% Novelty of our architecture
Different from previous approaches~\citep{LampleSRDJ19,fan_augmenting_2021,QAMAT}, our query representation is computed at an early transformer layer, whereas retrieved key and value embeddings are incorporated into the model at a later layer. 
This design only requires one forward pass through the transformer model,
and allows memory retrieval to run concurrently with the transformer forward pass, and hence reduces the computational overhead (see \cref{fig:arch} for our architecture).

%
% Our method differs from previous works~\citep{MemNet,E2EMemNet,LampleSRDJ19,fan_augmenting_2021} in two aspects:
% %
% \begin{inparaenum}[\itshape 1\upshape)]
% %
% \item{we use Maximum Inner Product Search~(MIPS) to query relevant memory slots instead of exact softmax:
% MIPS is computationally efficient and scalable, and MIPS implementations such as \texttt{faiss}~\citep{faiss} enable searching across millions of vectors in milliseconds on a CPU;}
%
% \item{key and value representations are computed at different layers. The query representation is computed at an early transformer layer, whereas the value embeddings are incorporated into the model at a later layer. This allows MIPS search to run concurrently with the transformer forward pass, and hence reduces the computational overhead.}
% %
% \end{inparaenum}

% So we introduce a middle ground between closed-book QA models and retrieve-and-read models, that can incorporate external knowledge into transformers without slowing down the model too much.

% Our pre-training
With this architecture, it is important that the key-value memory accurately represent the knowledge source, and the transformer learns a strategy to incorporate the retrieved key-value representations into the model.
Therefore, we introduce pre-training tasks~(\cref{sec:pretrain}), which include auto-encoding objectives to represent the questions and answers, and a question answering task to learn an implicit strategy to incorporate multiple key-value memory slots.
% that learn to ensure that our key-value memory encodes the knowledge source properly using auto-encoding task, and we learn an implicit strategy to incorporate multiple key-value memory slots by pre-training it with a question-answering task on PAQ.
Our ablation study (\cref{sec:ablation}) shows that our pre-training tasks are crucial for the performance, and removing any of them could lead to more than 10pp drop in EM score on ODQA datasets.

% Contributions
Our contribution can be summarised as follows:
\begin{inparaenum}[\itshape i\upshape)]
\item{we introduce \ModelName{}, an efficient memory access module to augment the transformer architecture;}
\item{we exploit PAQ as our knowledge source, and propose pre-training objectives to encode QA-pairs as key-value memory and to learn integration strategy to incorporate multiple memory slots into transformers;}
\item{experimental results on various knowledge-intensive tasks show that our proposed method significantly outperforms vanilla transformer baselines, while retaining a similar inference speed.}
\end{inparaenum}

% 1. Efficient extension of transformers
% 2. Scalability to large scale knowledge
% 3. Interpretability
% 4. Dense representation of knowledge

\section{Related Work}

\paragraph{Retrieve-and-Read Models for ODQA}
% The goal of Open-domain Question Answering (\acrshort{ODQA}) is to answer questions without
Open-domain question answering is a task that aims to answer a open-domain question without given context passages. 
% This task requires systems to either access external corpus or knowledge base, or memorise facts internally. 
Many \acrshort{ODQA} systems follow a two-steps \emph{retrieve-and-read} architecture~\citep{DBLP:conf/acl/ChenFWB17} where, in the first step, a \emph{retriever} model collects a set of relevant passages, and then a \emph{reader} model processes the retrieved passages and produces the answer~\citep{DBLP:conf/emnlp/MinCHZ19,DBLP:conf/naacl/YangXLLTXLL19,DBLP:conf/emnlp/WangNMNX19,DPR,REALM, RAG,FiD}.
%
% They could be further categorised into \emph{extractive ODQA models}~\citep{DBLP:conf/emnlp/MinCHZ19, DBLP:conf/naacl/YangXLLTXLL19, DBLP:conf/emnlp/WangNMNX19, DPR} and \emph{generative ODQA models}~\citep{RAG,FiD}.
% 
%Despite that retrieve-and-read systems achieve relatively high accuracy, they have to encode a large chunk of retrieved text together with the question, which leads to high computational cost and slow inference. 
%
Despite their high accuracy, retrieve-and-read systems have a high computational footprint, since they need to process a potentially large number of passages~\citep{DBLP:conf/emnlp/WuRMS20,DBLP:conf/acl/WuMS020}.

\paragraph{Efficient OQDA Systems}

One simple approach to accelerate \acrshort{ODQA} is \emph{Closed-Book QA} (CBQA) -- a sequence-to-sequence model~\citep{DBLP:conf/nips/SutskeverVL14,DBLP:conf/acl/KalchbrennerGB14} such as T5~\citep{T5} or BART~\citep{BART} is fine-tuned on ODQA data, by training it to produce the answer given the question.
CBQA models are substantially faster than retrieve-and-read approaches.
However, since they solely rely on their parameters to store factual knowledge, their capacity is limited by the model size, and hence they often produce less accurate results than retrieve-and-read methods~\citep{DBLP:conf/eacl/LewisSR21,DBLP:journals/corr/abs-2109-01156}.
Another efficient approach is retrieving semantically similar questions from a large collection of QA pair and returning the corresponding answers.
\citet{paq} propose PAQ, a 65 million QA dataset that is constructed with the objective to cover the most probably-asked questions in Wikipedia.
RePAQ~\citep{paq}, a retrieval-based QA system built on PAQ, won the EfficientQA competition~\citep{EfficientQA} in 2020, outperforming CBQA models by a large margin. 
In this work, we choose PAQ as our knowledge source, but different from RePAQ, we develop a generative model. Our results show that \ModelName outperforms RePAQ while matching its efficiency.

\paragraph{Memory-Augmented Transformers}
\citet{geva_transformer_2021} show that the Feed-Forward Network (FFN) layers in Transformer-based language models behave similarly to like key-value memories, where keys capture input patterns, and values map to the output vocabulary.
Based on this finding, \citet{Kformer} propose to extend the FFN layers by concatenating a dense representation of the corpus to the layer weights. % the encoded knowledge to weight matrix of the linear layer. 
\citet{fan_augmenting_2021} introduce %KIF 
a neural module to access a fixed external memory, showing that it can lead to significant improvements on downstream generative dialogue modelling tasks.
Concurrently to our work, \citet{QAMAT} propose QAMAT, a method to augment Transformer layers with a key-value memory network encoding question-answer pairs.
%
%QAMAT requires two passes through the encoders, one pass to retrieve memory values and another pass to concatenate memory into the input. In contrast, our proposed method only needs one forward pass through the encoder, and hence is more efficient. Empirically, we show that our method is 5 times faster than QAMAT, even when ours runs on smaller hardware (one A100 GPU compared to QAMAT's 32 TPU-v3).
%
QAMAT requires two inference steps through the encoder: one to retrieve memory values, and another for concatenating the retrieved values to the input.
In contrast, our proposed method only requires a single inference steps, resulting in a significantly smaller computational footprint.
Empirically, we show that our method is $\approx 5$ times faster than QAMAT, even when using fewer hardware resources.
% (one A100 GPU compared to QAMAT's 32 TPU-v3).

%\section{Proposed Method}
\section{Efficient Memory-Augmented Transformer}  \label{sec:emat}
In this work we propose Efficient Memory-Augmented Transformer (\ModelName), %to be capable of generating answers enhanced by retrieving from a key-value memory with millions of question-answer pairs.
a model architecture that uses a key-value memory to store millions of dense question-answer representations to inform its predictions (see \cref{fig:arch}).
Given an input sequence $X=(x_1, \cdots, x_{|X|})$, \ModelName's encoder first produces a dense query $\mathbf{q}$ to retrieve from the memory $\mathbf{M}$.
The returned key-value representations corresponding to the retrieved $k$ key-value pairs are $Z = (z_1, \cdots, z_k)$.
Finally, the decoder generates the target sequence $Y=(y_1, \cdots, y_{|Y|})$ conditioned on the input $X$ and retrieved key-value pairs $Z$.
%
% The training objective can be formulated as maximising the probability of generating the target sequence $Y$ given the input $X$:
%
% \begin{equation*}
% P(Y \mid X) = \mathbb{E}_{Z\sim P_{\eta}(\cdot \mid X, \mathbf{M})} [P_{\theta}(Y \mid X, Z)],
% \end{equation*}
% %
% where $\eta$ and $\theta$ denote the parameters of the retriever and the generator, respectively. 
% %

%
% In this section, we first describe the and key-value memory and retrieval method in \cref{sec:kvm} and \cref{sec:retriever}, and then introduce key-value integration method in \cref{sec:integration}.
% %
% Finally, we describe the training and inference algorithms \cref{sec:training} and \cref{sec:inference}.
%

%
\subsection{Key-Value Memory} \label{sec:kvm}
%
% In this work, to populate the key-value memory module $\mathbf{M}$, we use PAQ~\cite{paq}, a set of $65$ million generated QA pairs.
% %
% Here $\mathbf{M}$ contains the representations of all QA pairs in PAQ, and \ModelName uses MIPS to retrieve the most relevant QA pair representations from $\mathbf{M}$.

%
%
The key-value memory $\mathbf{M}=(\mathbf{K}, \mathbf{V})$ contains representations of keys $\mathbf{K}$ and values $\mathbf{V}$, with each key $\mathbf{k}_i$ mapping to one value $\mathbf{v}_i$.
Since we use PAQ~\citep{paq} as our knowledge source, each key represents a question, and its value represents the corresponding answer.
We use \ModelName's encoder to encode the question and the answer separately, and it produces key and value embeddings from $l_k$-th and $l_v$-th layer of encoder respectively.
\cref{fig:arch} (left) shows details regarding how key (question) and value (answer) are encoded in \ModelName.
To encode the key embeddings, we first concatenate a prefix $\prefix$ of length $\textup{P}$ with the question $q$ as input, and then obtain the hidden states at the $l_k$-th layer $\mathbf{h}^{l_k} = [\mathbf{h}_1^{l_k}, \cdots, \mathbf{h}_n^{l_k} ]$, where $n$ is the length of the question $q$ prepended with \prefix.
Then, $\mathbf{h}^{l_k}$ is passed through a convolutional neural network layer to produce $[\mathbf{c}_1, \cdots, \mathbf{c}_{n}]$, 
and we use the prefix part as our final key representation $\mathbf{k} =[\mathbf{c}_1, \cdots, \mathbf{c}_{\textup{P}}] \in \mathbb{R}^{\textup{P} \times h}$ .
%
% Finally, the key $k$ is obtained by reshaping $k'$ into $\textup{P}$ $h$-dimensional embeddings, \ie $\mathbf{k} = [{k'}_1, \cdots, {k'}_{\textup{P}}]$.
%
% To facilitate dense retrieval, we build the index $k$ for each key-value pair by averaging the key embeddings: $k = \frac{1}{\textup{P}} \sum_{i=1}^{\textup{P}}{k'}_{i}$.
% 
For value embeddings, we prepend a prefix to the answer, feed $[\prefix; a]$ into the model,
%
%The prefix will interact with the original input via self-attention to gather the input's information when forwarding in Transformer layers.
%
and use the prefix's representation at the $l_v$-th layer of encoder $\mathbf{v} = [\mathbf{h}_1^{l_v}, \cdots, \mathbf{h}_{\textup{P}}^{l_v}] \in \mathbb{R}^{\textup{P} \times h}$ as our value representation, where $h$ is the size of hidden representations.
% , where $\textup{P}$ is the prefix length and $h \in \mathbb{R}^h$. It aggregates the answer's information by $l_v$ layers transformers.

%
\subsection{Memory Retrieval} \label{sec:retriever}
The goal of the retriever is to retrieve relevant entries from the key-value memory $\mathbf{M}$ to inform the downstream generation tasks.
\ModelName's encoder embeds the question into a query $\mathbf{q}$ %with the same procedures to obtain the index $\mathbf{k}$ in $l_k$ layer.
using the same procedure as the key embeddings, described in \cref{sec:kvm}.
We conduct an extra step of flattening for both $\mathbf{q}$ and $\mathbf{k}$ by averaging:   
$\bar{\mathbf{k}} = \text{flatten}(\mathbf{k}) = \frac{1}{\textup{P}} \sum_{j=1}^{\textup{P}} \mathbf{k}_{j}$.
The key-value encoder shares the parameters with the question encoder,
and we define the query-key similarity by the inner product between the flattened query representation and key representation $\text{sim}(\mathbf{q}, \mathbf{k}) = \langle \bar{\mathbf{q}}, \bar{\mathbf{k}} \rangle$.
%, making the retrieval symmetric by question.
%It adopts the dense Maximum Inner Product Search (MIPS) paradigm to retrieve key-value pairs.
%
At inference time, this operation can be efficiently computed using Maximum Inner Product Search (MIPS)
to retrieve the top-$k$ key-value pairs $Z = \{(\mathbf{k}_i, \mathbf{v}_i)\}_{i=1}^{k}$ based on the similarity.
MIPS implementations such as \texttt{faiss}~\citep{faiss} enable searching across millions of vectors in milliseconds on a CPU.
%
%Denote the retrieved top-$k$ key-value pairs in descending order.
%
%We list the retrieved top-$k$ key-value pairs $Z = \{(\mathbf{k}_i, \mathbf{v}_i)\}_{i=1}^{k}$ in decreasing order, and integrate them in later layers of \ModelName encoder.
%
The retrieved key-value pairs $Z$ are then integrated in later layers of \ModelName's encoder.
\subsection{Key-Value Integration} \label{sec:integration}
Once we have retrieves the top-$k$ key-value pairs $Z$, they need to be incorporated into the model. 
%
% The retrieval performs in the $l_k$-th layer of encoder, we then integrate the retrieved key-value pairs $Z$ start from $l_c$-th layer.
%
%
More specifically, in the $l_c$-th layer, all the key embeddings in $Z$ are ordered by their corresponding similarity scores, and concatenated into a matrix $\mathbf{K}'=[\mathbf{k}_i, \cdots, \mathbf{k}_k] \in \mathbb{R}^{\textup{P} k \times h}$.
Then it is prepended to the $l_c$-th layer's hidden states.
%
% Then, such hidden states and the prepended keys interact with each other via self-attention until $l_v$-th layer.
%
To distinguish the different keys, we additionally add relative positional encodings to $\mathbf{K}'$.
% the representations produced by the $l_c$-th layer.
%
In the $l_v$-th layer, the value embedding in $\mathbf{Z}$ are concatenated in the same way to produce $\mathbf{V}'$, and it is \emph{added} to the positions where their corresponding key embeddings are prepended to.
The updated hidden states continue the forward pass of the remaining transformer encoder layers.
%  and outputs the hidden states that integrates the retrieved key-value pairs.
%
% Here, we first concatenate keys and add the values in a later layer. 
% The pre-concatenate keys are as placeholders, informing the model how to use the upcoming value in advance.
%
%
Finally, the decoder generates the answer condition on the output of the encoder, which already integrates the retrieved key-value representations.

\section{Training Pipeline of \ModelName} \label{sec:pipeline}

\subsection{Pre-Training} \label{sec:pretrain}
% To obtain better representation of key and value, we pre-train \ModelName with auto-encoding task and answer generation task.

\paragraph{Auto-encoding Tasks}
We use T5-base's pre-trained parameters to initialise \ModelName, but the prefix embeddings and key encoder's convolutional layer are trained from scratch.
%
%Thus, the prefix is difficult to learn to gather the information from input at the beginning, and the key is difficult to learn an expected representation for retrieving.
%
To obtain better representation of key and value, we pre-train \ModelName with auto-encoding training objectives. % auto-encoding task.
We use PAQ-L1, a simplified version of PAQ that consists of 14M QA pairs, as the pre-training corpus.
The model is trained to recover the input question $x$ given the key embeddings $\mathbf{k}$, and the answer $y$ given the value embeddings $\mathbf{v}$, as shown in \cref{fig:arch} (left).
The two tasks key auto-encoding (KAE) and value auto-encoding (VAE) can be formalised as:
\begin{equation*}
\begin{aligned}
\mathcal{L}_{\text{KAE}} = & - \sum_{i=1}^{|X|}  \log P(x_i \mid \mathbf{k}, x_{<i}), \\
\mathcal{L}_{\text{VAE}} = & - \sum_{i=1}^{|Y|}  \log P(y_i \mid \mathbf{v}, y_{<i}).
\end{aligned}
\end{equation*}

\paragraph{Generation Task}
Besides the problem of representing questions and answers in key-value memory $\mathbf{M}$, we also need the model to make use of $\mathbf{M}$ for downstream tasks. %the original T5 model is not capable of using dense key-value pairs to generate text.
Thus, it is also critical to pre-train the model to learn the key-value integration module defined in \cref{sec:integration}.
Since PAQ provides a large number of QA pairs, we consider a generation task built on PAQ to pre-train the model.
More concretely, for each QA pair $(x, y)$ in PAQ, we use the RePAQ model~\citep{paq} to retrieve $10$ other relevant QA pairs from PAQ, and retrieve their corresponding keys $\mathbf{K}'_x = [\mathbf{k}_{1}, \cdots, \mathbf{k}_{10}]$ and values $\mathbf{V}'_x = [\mathbf{v}_{1}, \cdots, \mathbf{v}_{10}]$ from the memory $\mathbf{M}$.
Then, the model is trained to generate the answer $y$ given the question $x$ and the key-value embeddings corresponding to the retrieved QA pairs.
% :
% \begin{equation*}
% \mathcal{L}_{gen} = -\sum_{i=1}^{|Y|}  \log p(y_i | X, Z, y_{<i})
% \end{equation*}
%
% The overall objective of pre-training is formalised as:
% \begin{equation*}
% \mathcal{L}_{pre} = \lambda_{kae} \mathcal{L}_{kae} + \lambda_{vae} \mathcal{L}_{vae} + \lambda_{gen} \mathcal{L}_{gen}
% \end{equation*}
%
The objective can be defined as follows:
\begin{equation*}
\mathcal{L}_{\text{Gen}} = - \sum_{i=1}^{|Y|}  \log P(y_i \mid x, \mathbf{K}'_x, \mathbf{V}'_x, y_{<i}).
\end{equation*}
We adopt a multi-task pre-training objective to minimise $\mathcal{L}_{\text{KAE}} + \mathcal{L}_{\text{VAE}} + \mathcal{L}_{\text{Gen}}$.

\subsection{Fine-Tuning on Downstream Tasks} \label{sec:finetune}

After pre-training, we fine-tune both the memory retrieval module and the generation of \ModelName on the downstream tasks.
\paragraph{Retrieval Objective}
Learning to retrieve relevant key-value pairs that provide useful evidence to solve a given task can be challenging due to the lack of labelled data.
To solve this problem, we propose a weakly-supervised method to optimise the retriever.
Specifically, we first rank all retrieved key-value pairs retrieved from the memory $\mathbf{M}$ by their inner product scores. % using the up-to-date model.
%
%We first consider the top-$n$ key-value pairs are comparatively semantic-related to the input text, then we sample a lexical-matched pairs from them to optimise the retriever.
%
%For QA tasks, the lexical-matched pairs containing the same answer to the target answer.
%
We consider the top retrieved key-value pairs: for each retrieved key-value pair, if its corresponding answer is lexically matched with the target output, then the key-value pair is selected as positive sample to optimise the retriever.
For short output generation tasks such as ODQA, we match the answer corresponding to the retrieved value with the target answer.
For long sequence generation tasks such as open-domain dialogue and long-form QA, we normalise the target sequence (\ie, lower-casing and removing stop words), and check whether if the retrieved value (answer) is contained in the normalised target sequence.
%
%Since these semantic-related and lexical-matched pairs are more likely to give the correct answer directly, they can be used to provide a weakly-supervised training to the retriever.
%
Since these key-value pairs are more likely to lead to the correct answer, they can be used to provide a weakly-supervised training signal to the retrieval component of \ModelName.
We denote the selected positive key-value pairs as $Z^+ = (z_1^+, \cdots, z_r^+)$, % and the corresponding indices as $(k_1^+, \cdots, k_r^+)$.
where each pair $z_{i}^{+} = (\mathbf{k}_{i}^{+}, \mathbf{v}_{i}^{+})$ is composed by a key component $k_{i}^{+}$ and a value component $v_{i}^{+}$.
We sample a key-value pair $z_{i}^+$ from $Z^+$ based on the similarity between the corresponding key $\mathbf{k}_{i}^{+}$ and the query $\mathbf{q}$:
\begin{equation*}
\begin{aligned}
P_{\eta}(z_i^+ \mid q) & = \frac{\exp ( \text{sim}(\mathbf{q}, \mathbf{k}_i^+  ))} {\sum_{j=1}^r \exp ( \text{sim}( \mathbf{q}, \mathbf{k}_j^+ ) ) }, \\
z^+ & \sim P_\eta (\cdot \mid q, Z^+).
\end{aligned}
\end{equation*}
We then select $m$ negative pairs $\{z^-_j\}_{j=1}^m$ that do not match the target sequence.
Finally, the positive pairs $z^+$ and the negative pairs $z^-$ are used to train the retriever, by optimising the following objective:
\begin{equation*}
\begin{aligned}
&\mathcal{L}_{\text{Ret}} =
\\
&-\log \frac{\exp (\text{sim}(\mathbf{q}, \mathbf{k}_i^+  )  )}{ \exp ( \text{sim}(\mathbf{q}, \mathbf{k}_i^+  ) ) + \sum_{j=1}^{m} \exp( \text{sim}(\mathbf{q}, \mathbf{k}_j^-  ) ) }.
\end{aligned}
\end{equation*}
%
%
% The retrieval training method can be seen as an instance of the %is also an 
% EM algorithm: the model estimates the target key-value pairs by sampling from $Z^+$ (Expectation step), and updates the parameters using $z^+$ as optimisation objective (Maximisation step).
% %
% We show the effectiveness of this approach in our experiments (\cref{sec:experiments}).
%

%
%\paragraph{Local Memory}
\paragraph{Memory Caching for More Efficient Training}
As described above, %we should rank the entire memory to select a semantic-related and lexical-matched retrieval target.
\ModelName uses MIPS for retrieving the key-value pairs that are the most relevant to solve the current task.
%
%It is costly when the memory is very large, because we should re-encode the memory and rank the memory using the up-to-date model in every batch iterations.
%
However, updating the memory $\mathbf{M}$ after each training update may not be feasible when the number of entries in $\mathbf{M}$ is very large.
To alleviate this problem, we design a \emph{memory caching} mechanism. %to cache a small part of memory as coarse searching space.
At the beginning of each training epoch, we freeze the memory $\mathbf{M}$ and, for each training example, we retrieve the top-$n$ key-value pairs.
%
%for each training example, we retrieve the top-$n$ key-value pairs from $\mathbf{M}$ to build a local memory cache $\mathbf{M}_{l}$, which is then frozen for the whole epoch.
%
%The local memory cache is frozen for the whole epoch, and the model only encode the local memory to select retrieval target in a batch iteration.
%
The memory $\mathbf{M}$ is updated only at the end of the epoch by re-encoding all entries in the knowledge source.
%

% Local memory mechanism is similar to a recall-and-rank architecture. $\mathbf{M}_{loc}$ is used as a cache storing the coarse key-value pairs, and the model reranks $\mathbf{M}_{loc}$ to select fine-grained retrieval targets. Updating local memory asynchronously allows us to update the model and key-value pairs in an end-to-end manner.

%
\paragraph{Overall Fine-Tuning Objective}
The generator is optimised to generate the target $y$ given the input $x$ and the top-$n$ retrieved key-value pairs $Z$:
\begin{equation*}
\mathcal{L}_{\text{Gen}} = -\sum_{i=1}^{|Y|}  \log P(y_i \mid x, Z, y_{<i}),
\end{equation*}
% \begin{equation}
% \mathcal{L}_{gen} = -\sum_{i=1}^{|\mathcal{R}|}  \log p(s_i | X, s_{<i}) ,
% \end{equation}
so the overall fine-tuning objective is $\mathcal{L}_{\text{Ret}} + \mathcal{L}_{\text{Gen}}$.

\subsection{Inference} \label{sec:inference}

During inference, we use a fast Hierarchical Navigable Small World~\citep[HNSW,][]{DBLP:journals/pami/MalkovY20} graph index, generated by \texttt{faiss}, to search and retrieve from the key-value memory $\mathbf{M}$.
If the $l_k < l_c$, the search process can run in parallel with the evaluation of the layers $l_k + 1, \cdots, l_c - 1$ in \ModelName.
%
%Because the search performs at $l_k$-th layers of encoder in CPU, it does not conflict with the encoder's forwarding in GPU, and the encoder await the retrieval results at $l_c$-th layers.
%
Since the search process can be efficiently executed on CPU, it does not increase the GPU memory requirements of the model.

\section{Experiments} \label{sec:experiments}

\subsection{Experimental Setup}

\paragraph{Datasets}
We consider several knowledge-intensive NLP tasks, including Open-Domain Question Answering (ODQA), Open-Domain Dialogue (ODD), and Long-Form Question Answering (LFQA).
% ODQA
In ODQA, the aim is to answer factual questions %without context
using a large collection of documents of diversified topics. %, and the answer is usually a short piece of text such as an entity.
We choose three commonly used datasets -- NaturalQuestions~\citep[NQ,][]{NaturalQuestions}, TriviaQA~\citep[TQA,][]{TriviaQA}, and WebQuestions~\citep{WebQuestions}.
In addition, we consider two generation tasks from the Knowledge Intensive Language Tasks~\citep[KILT,][]{KILT} benchmark to test whether our method generalises to tasks beyond ODQA.
% WoW
Specifically, we consider Wizard-of-Wikipedia~\citep[WoW,][]{WoW} for ODD.
This task requires modelling long dialogue history and acquire relevant Wikipedia knowledge to produce a response utterance.
% ELI5
Furthermore, we consider the Explain Like I’m Five~\citep[ELI5,][]{ELI5} dataset for LFQA.
In ELI5, answers are often more diverse and open-ended compared to ODQA, and they tend to be significantly longer -- they can be composed by several sentences.
\paragraph{Knowledge Source}
We use PAQ~\citep{paq} as our knowledge source, and encode question-answer pairs in the model's key-value memory. 
% 
% PAQ is a collection of 65 million QA pairs generated from Wikipedia articles, with the goal to cover the most likely asked questions in ODQA.
%
Since the generative model used to generate the QA pairs in PAQ was trained on NaturalQuestions and TriviaQA, PAQ has a high coverage for these two ODQA datasets.
In this work, we also evaluate on tasks beyond ODQA, where it is not clear how PAQ can be used. 
Therefore, our evaluation on ODD and LFQA aims to demonstrate that \ModelName generalises to different knowledge-intensive generation tasks using PAQ as the underlying knowledge source.
%
%\yw{Defend ourselves a bit that "we could have used other knowledge sources, such as KB, but leave it to future work"?}
%

%
\paragraph{Baselines}
We compare our method with three types of baselines: \emph{parametric models}, \emph{retrieval-only approaches}, and \emph{retrieval-augmented models}.
Parametric models fine-tune sequence-to-sequence PLMs such as T5~\citep{T5} or BART~\citep{BART} on a datasets, %by treating the questions or dialogue history as input sequence, and produce an output sequence as answer or response respectively.
by casting each task as a sequence generation problem conditioned on the input.
%
% We test parametric models with different model sizes as well.
%
In our experiments, we consider parametric models of multiple sizes, including T5-base, T5-large, T5-3B, T5-11B~\citep{DBLP:conf/emnlp/RobertsRS20}, and BART-large~\citep{BART}.
Retrieval-only approaches retrieve the most relevant information from the knowledge source (PAQ), and return the top answer as output.
In ODQA benchmark we use the RePAQ model proposed by~\citet{paq}; in ODD and LFQA, we use the \ModelName key retrieval module described in \cref{sec:retriever} as the retriever.
Retrieval-augmented models such as RAG~\citep{RAG} or FiD~\citep{FiD} retrieve relevant passages from Wikipedia using a dense retriever such as DPR~\citep{DPR}, and then use the retrieved passages and the input sequence to condition the generation process.

\paragraph{Pre-Training and Fine-Tuning Configurations}
We base our \ModelName on T5~\citep{T5}, and initialise our model with the pre-trained parameters from T5-base.\footnote{We use the \href{https://huggingface.co/t5-base}{original version of T5 without SSM} to ensure that our results are comparable with the baselines.}
To evaluate the speed and accuracy of our proposed method under different computation environments, we pre-train and fine-tune \ModelName using two settings.
In the former setting, we set $l_k=3$, $l_c=3$, $l_v=7$, which emulates an environment where key embeddings has fast access, but there is delay in acquiring value embeddings; we refer to this setting as \emph{Fast Key, Slow Value} (FKSV).
In the latter setting, $l_k=3$, $l_c=10$, $l_v=11$, which corresponds to a scenario where both key querying and value reading can have significant delays.
We refer to this setting as \emph{Slow Key, Slow Value} (SKSV).
All details on the training hyperparameters the hardware used in our experiments are available in \cref{sec:hyper}.

\subsection{Open-Domain Question Answering}

\newcommand{\cbqa}{\textbf{Parametric models}}
\newcommand{\retqa}{\textbf{Retrieval-only models}}
\newcommand{\obqa}{\textbf{Retrieval-augmented models}}

\begin{table}
\centering
\resizebox{\columnwidth}{!}{
    \begin{tabular}{lcccc}
    \toprule
    \multirow{2}{*}{\textbf{Model}} & \multicolumn{2}{c}{\bf{NQ}}  & \textbf{TQA} & \textbf{WQ}  \\ 
     & EM & Q/s & EM & EM  \\
    \midrule
    \multicolumn{2}{l}{\cbqa} & \\
    T5-base~\citep{DBLP:conf/emnlp/RobertsRS20} & 25.8 & 1600 & 24.4 & 26.6  \\
    T5-large~\citep{DBLP:conf/emnlp/RobertsRS20}   & 27.6 & 570  & 29.5 & 27.7 \\
    T5-3B~\citep{DBLP:conf/emnlp/RobertsRS20}      & 30.4 & 55 & 35.1 & 33.6  \\
    T5-11B~\citep{DBLP:conf/emnlp/RobertsRS20}     & 32.6 & - & 42.3 & 37.2  \\
    BART-large~\citep{BART} & 26.5 & 570 & 26.7 & 27.4  \\
    % BART-Large, pre-finetuned on PAQ & 32.7 & - & 33.2 & - \\
    % GPT-3~\citep{GPT3} & 14.6 & - & 64.3 & 14.4 \\
    % PaLM~\citep{PaLM} & 24.7 & - & 71.3 & 19.0 \\
    % Gopher~\citep{Gopher}  & 35.2 & - & 60.5 & - \\
    \midrule 
    \multicolumn{2}{l}{\retqa} &  \\
    Dense Retriever~\citep{DBLP:conf/eacl/LewisSR21} & 26.7 & - & 28.9 & - \\
    DensePhrases~\citep{DBLP:conf/acl/LeeSKC20} & 40.9 & 18 & 50.7 & -    \\
    RePAQ-base~\citep{paq} & 40.9 & 1400 & 39.7 & 29.4 \\
    RePAQ-large~\citep{paq} & 41.2 & 1100 & - & - \\
    RePAQ-xlarge~\citep{paq} & 41.5 & 800 & 41.3 & - \\
    \midrule 
    \multicolumn{2}{l}{\obqa} &  \\
    REALM~\citep{REALM}       & 40.4 & - & 55.8 & 40.7 \\
    DPR~\citep{DPR}           & 41.5 & 2.7 & 57.9 & 42.4 \\
    QAMAT~\citep{QAMAT}       & 44.7 & 240* & 48.0 & 39.4 \\
    RePAQ rerank~\citep{paq}  & 45.7 & 55 & 48.9 & 37.6 \\
    RAG~\citep{RAG}           & 44.5 & 9.6 & 56.8 & 45.2 \\
    FiD-base~\citep{FiD}      & 48.2 & 3.7 & 65.0 & 32.4 \\
    FiD-large~\citep{FiD}     & 51.4 & 1.4 & 67.6 & -    \\
    % RAG~\citep{RAG}           & 44.5 & 5 & 56.8 & 45.2 \\
    % FiD-base~\citep{FiD}      & 48.2 & 2 & 65.0 & 32.4 \\
    % FiD-large~\citep{FiD}     & 51.4 & 0.5 & 67.6 & -    \\
    % 11 & \retqa & -multitask (with reranker) & 47.6 & \underline{52.1}\\
    % 12 & QA-pair retriever & -multitask w/ FiD-Large Backoff & \textbf{52.3} & 67.3\\
    \midrule
    \multicolumn{2}{l}{\textbf{Ours}} &  \\
    % 16  & \retqa                & \ModelName{}-retriever    & 42.3 & 43.5 & 33.5 \\
    \ModelName-FKSV  & 44.3 & 1000 & 44.4 & 36.7 \\
    \ModelName-SKSV  & 43.3 & 1200 & 43.7 & 33.2 \\
    \bottomrule
    \end{tabular}
}
\caption{Exact Match (EM) results for %highest accuracy configurations 
\ModelName in comparison to recent state-of-the-art systems. $*$ QAMAT runs on 32 TPU-v3 with 1024GB TPU memory, whereas ours run on A100 GPU with 40GB GPU memory. 
}
\label{tab:odqa_results}
\end{table}

% results of T5-base/large are from "Switch transformers: Scaling to trillion parameter models with simple and efficient sparsity (W Fedus, B Zoph, N Shazeer)"

% result of BART-large-WebQuestion is from "Challenges in Generalization in Open Domain Question Answerin (Linqing Liu Patrick Lewis‡ Sebastian Riedel Pontus Stenetorp)"

% result of FiD-base, 100 docs-WebQuestion is from "End-to-End Training of Multi-Document Reader and Retriever for Open-Domain Question Answering (Devendra Singh Sachan, Siva Reddy, William Hamilton, Chris Dyer, Dani Yogatama)"

% results of RePAQ on WebQuestion are from "Augmenting Pre-trained Language Models with QA-Memory for Open-Domain Question Answering (Wenhu Chen, Pat Verga, Michiel de Jong, John Wieting, William W. Cohen)"

\cref{tab:odqa_results} shows the experimental results on three ODQA datasets: NQ~\citep{NaturalQuestions}, TQA~\citep{TriviaQA}, and WQ~\citep{WebQuestions}.
We report the Exact Match (EM) scores and the average inference speed measured by queries per second (Q/s).
% 
% Our proposed \ModelName (FKSV) achieves 44.3\% EM score on NaturalQuestions and a query speed of 1000 Q/s. 
% 
Compared with \emph{parametric models}, our proposed method yields substantially higher EM scores across three datasets.
\ModelName-FKSV outperforms T5-base, which share the same backbone model, by 18.5, 20.0, 10.1 percentage points on NQ, TQA and WQ respectively.
These results indicate that our method of augmenting transformer with key-value memory effectively extends model's knowledge capacity.
Compared with \emph{retrieval-only models}, our method also demonstrates strong performance. 
RePAQ baselines also exploit PAQ as knowledge source, and hence is comparable with our method.
Our \ModelName-FKSV outperforms the best RePAQ retriever (RePAQ-large) by 2.8 and 3.1 percentage points on NQ and TQA respectively.
% \ModelName outperforms RePAQ-base, RePAQ-large, and RePAQ-xlarge, while retaining over 1000 Q/s inference speed.
Speed-wise, \ModelName can answer 1000-1200 Q/s, which is a high throughput in ODQA and is comparable to some of the fastest parametric models and retrieval-only models.
In ODQA, \emph{retrieval-augmented models} are known to be highly accurate, but are also computationally inefficient~\citep{EfficientQA}.
\ModelName is significantly faster than these models.
For example, FiD-base uses the same backbone T5-base model as \ModelName, but retrieves and concatenates 20 to 100 passages from Wikipedia.
Despite being less accurate on NQ and TQA, \ModelName is two orders of magnitude faster than FiD-base and more accurate on WQ.
On NaturalQuestions in terms of EM score, our method outperforms REALM and DPR, and is comparable with QAMAT and RAG.
QAMAT~\citep{QAMAT} is a concurrent work to ours and is the fastest among the retrieval-augmented models. 
But QAMAT runs on 32 TPU-v3~\citep{TPU}, which have roughly 1024GB TPU memory, and the MIPS search is conducted on TPU. 
In contrast, \ModelName runs on a single A100 GPU with 40GB GPU memory, and the MIPS search is executed on CPU.
Despite using substantially fewer resources, \ModelName-FKSV is roughly $4.2$ times faster than QAMAT, and \ModelName-SKSV is $5$ times faster.

\subsection{Generalisation to Open-Domain Dialogue and Long-Form QA}

\begin{table}
\centering
\resizebox{\columnwidth}{!}{
    \begin{tabular}{lccc}
    \toprule
    % \multirow{2}{*}{\textbf{Model}} & \multicolumn{3}{c}{\textbf{WoW}}  \\ 
    % \cmidrule(r){2-4}  \cmidrule(r){5-6}
    \textbf{Model} & \textbf{F1} & \textbf{R-L} & \textbf{U/s}  \\ 
    \midrule
    \textbf{\cbqa} \\
    Trans MemNet~\cite{WoW}   & 11.85 & 10.11 & -       \\ 
    BART-large~\citep{BART}      & 12.86 & 11.77 & 55    \\
    T5-base~\citep{T5} & 13.53 & 12.40 & 160 \\   % BLUE2=17.72
    \midrule
    \textbf{\obqa} \\
    BART + DPR~\citep{KILT}      & 15.19 & 13.23 & 0.7   \\
    RAG~\citep{RAG}  & 13.11 & 11.57 &  3.4   \\ 
    \midrule
    \textbf{\retqa} \\
    RePAQ w/ \ModelName{} key encoder & 1.84 & 1.48 & -  \\
    \midrule
    \textbf{Ours} \\
    \ModelName-FKSV   & \textbf{15.78} & \textbf{14.73} & 141    \\  % BLUE2=23.87
    \ModelName-SKSV   & 15.35 & 14.68 & \textbf{150}    \\ 
    \bottomrule
    \end{tabular}
}
    \caption{Results on the Wizard-of-Wikipedia dataset from the KILT benchmark.}
    \label{tab:wow_results}
\end{table}
% \footnotetext{KILT leaderboard: https://eval.ai/web/challenges/challenge-page/689/leaderboard}

% 27 tokens

\paragraph{Open-Domain Dialogue}
Open-domain dialogue is a dialogue task that requires accessing knowledge from Wikipedia to produce dialogue response.
\cref{tab:wow_results} shows the results on the open-domain dialogue dataset Wizard-of-Wikipedia~\citep{WoW} from the KILT~\citep{KILT} benchmark.
The utterances from dialogue history are concatenated into a input sequence, and the output sequence is the corresponding response utterance.
We follow~\citet{KILT} and evaluate the models with F1 and ROUGE-L metrics, and we also report the average number of utterances generated per second (U/s) for speed evaluation.

The results show that, our proposed \ModelName outperforms parametric models while retaining a similar inference speed.
\ModelName-FKSV outperforms T5-base by 2.25 F1 and 2.28 ROUGE-L points, while generating 141 utterances per second.
Surprisingly, \ModelName models also outperform retrieval-augmented models such as RAG and BART+DPR, which exploits Wikipedia as knowledge source.
It indicates that our method that encodes PAQ as key-value memory is capable of represent crucial information in Wikipedia, and generalises well to tasks beyond ODQA.
We also implement a RePAQ-equivalent retrieval-only model using \ModelName's key encoder. 
Since PAQ is a collection of QA pairs, simply retrieving relevant QA pairs for dialogue does not work well.
The large gap between \ModelName and RePAQ with \ModelName key encoder, together with the qualitative analysis in \cref{sec:qualitative}, demonstrates that \ModelName decoder does not simply copy information from the key-value memory, but exploits the key-value embeddings to generate novel response.

% \begin{table}[h]
% \renewcommand\arraystretch{1.2} 
% \setlength\tabcolsep{4pt}
% \centering
% \resizebox{0.48\textwidth}{!}{
%     \begin{tabular}{lccccc}
%     \toprule
%     \multirow{2}{*}{\textbf{Model}} & \multicolumn{3}{c}{\textbf{Wizard-of-Wikipedia}} & \multicolumn{2}{c}{\textbf{ELI5}} \\ 
%     \cmidrule(r){2-4}  \cmidrule(r){5-6}
%     & F1 & ROUGE-L & BLEU2$^\dag$ & F1 & ROUGE-L \\ \midrule
%     % \textbf{Model}  & \textbf{F1}    & \textbf{BLUE2} & \textbf{BLUE4} & \textbf{ROUGE-L} \\ \midrule
%     Trans MemNet    & 11.85 & 10.11 & -     & -     & -     \\ 
%     BART-large      & 12.86 & 11.77 & -     & 19.23 & 20.55 \\
%     T5-base         & 13.53 & 12.40 & 17.72 & 16.01 & 19.08 \\ \midrule
%     BART + DPR      & 15.19 & 13.23 & -     & 17.88 & 17.41 \\
%     RAG             & 13.11 & 11.57 & -     & 14.51 & 14.05 \\ \midrule
%     \ModelName{}    & 15.78 & 14.73 & 23.87 & 18.08     & 20.07     \\
%     \ModelName{}, retriever only & 1.84 & 1.48 & - & 1.40 & 1.65 \\
%     \bottomrule
%     \end{tabular}
% }
%     \caption{Results on Wizard-of-Wikipedia and ELI5, KILT leaderboard\protect\footnotemark . BLEU2$^\dag$ evaluated on validation set.}
%     \label{tab:wow_results}
% \end{table}
% \footnotetext{KILT leaderboard: https://eval.ai/web/challenges/challenge-page/689/leaderboard}

\begin{table}
\centering
\resizebox{\columnwidth}{!}{
    \begin{tabular}{lccc}
    \toprule
    \textbf{Model} & \textbf{F1} & \textbf{R-L} & \textbf{Q/s}  \\ 
    \midrule
    \textbf{\cbqa} \\
    BART-large~\citep{BART}      & 19.23 & 20.55 & 30    \\
    T5-base~\citep{T5} & 16.01 & 19.08 & 76 \\ 
    \midrule
    \textbf{\obqa} \\
    BART + DPR~\citep{KILT}      & 17.88 & 17.41 & 0.2   \\
    RAG~\citep{RAG}  & 14.51 & 14.05 &  0.4   \\ 
    \midrule
    \textbf{\retqa} \\
    RePAQ w/ \ModelName{} key encoder & 1.40 & 1.65 & -  \\
    \midrule
    \textbf{Ours} \\
    \ModelName-FKSV   & 18.42     & 20.61  & 67    \\
    \ModelName-SKSV & \textbf{19.03} & \textbf{20.91} & \textbf{71}    \\ 
    \bottomrule
    \end{tabular}
}
    \caption{Results on the ELI5 dataset from the KILT benchmark.}
    \label{tab:eli5_results}
\end{table}
% \footnotetext{KILT leaderboard: https://eval.ai/web/challenges/challenge-page/689/leaderboard}

% 186 tokens

% ---------------RL-------F1-
%               20.07   18.08
%----------------------------
%           	21.59	16.16	
% 	            21.06	17.98
%              	20.61	18.42
% 
%-----------------------------
% BART-large    20.55   19.23 

\paragraph{Long-Form Question Answering}
Results on the LFQA task ELI5~\citep{ELI5} (shown in \cref{tab:eli5_results}) reveals similar conclusions as in WoW.
Both \ModelName-FKSV and \ModelName-SKSV outperform the T5-base baseline by a large margin, 2.41pp and 3.01pp in F1, while retaining an inference speed to 67 Q/s and 71 Q/s, respectively.
\ModelName is both faster and more accurate than retrieval-augmented models on ELI5 too. 
Compared to RAG, \ModelName-SKSV is 4.52pp better in F1, 6.86pp better in ROUGE-L, and more than 160 times faster in inference speed.

% \yw{Should we discuss the differences between the two variants FKSV and SKSV?}

% \subsection{Speed and Asynchronous}
% \input{tables/speed_results}

\section{Analysis}

\subsection{Ablation Study} \label{sec:ablation}

\begin{table}
\centering
\resizebox{\columnwidth}{!}{
    \begin{tabular}{lccc}
    \toprule
    \textbf{Model}           & \textbf{NQ}   & \textbf{TQA}  & \textbf{WQ}   \\ \hline
    % v   T5-base~\citep{DBLP:conf/emnlp/RobertsRS20} & 25.8  & 24.4 & 26.6  \\
    % T5-large~\citep{DBLP:conf/emnlp/RobertsRS20}   & 27.6   & 29.5 & 27.7 \\
    % T5-3B~\citep{DBLP:conf/emnlp/RobertsRS20}      & 30.4  & 35.1 & 33.6  \\
    % T5-11B~\citep{DBLP:conf/emnlp/RobertsRS20}     & 32.6  & 42.3 & 37.2  \\
    \ModelName-FKSV  & 44.3 & 44.4 & 36.7 \\
    \hspace{2em} $-$ fine-tune & 30.6 & 32.4 & 25.6 \\ 
     \hspace{2em} $-$ auto-encoding tasks  & 28.5 & 34.6 & 12.9 \\
    \hspace{2em} $-$ generation task  & 28.7 & 24.7 & 31.4 \\
    \hspace{2em} $-$ all pre-training tasks & 27.1 & 17.7 & 6.0  \\ 
    \bottomrule
    \end{tabular}
}
    \caption{Ablation on the pre-training steps used by \ModelName, described in \cref{sec:pretrain}, measured using EM on NQ, TQA, and WQ: we analyse the impact of removing auto-encoding, generation, and all pre-training tasks from \ModelName's pre-training phase.}
    \label{tab:pretrain_results}
\end{table}

% \begin{table}
% \renewcommand\arraystretch{1.2} 
% \setlength\tabcolsep{4pt}
% \centering
% \resizebox{0.48\textwidth}{!}{
%     \begin{tabular}{lccc}
%     \toprule
%     \textbf{Model}           & \textbf{NQ}   & \textbf{TQA}  & \textbf{WQ}   \\ \hline
%     \ModelName-retriever, zero-shot     & 37.0 & 23.2 & 18.9 \\
%     \ModelName, zero-shot               & 30.6 & 32.4 & 25.6 \\ \hline
%     \ModelName-retriever, w/o pre-train  & 27.0 & 34.4 & 26.0 \\
%     \ModelName, w/o pre-train            & 38.1 & 17.7 & 6.0  \\ \hline
%     \ModelName-retriever, w/o auto-encoding task   & 36.9 & 38.1 & 28.6 \\
%     \ModelName, w/o auto-encoding task             & 28.5 & 34.6 & 12.9 \\ \hline
%     \ModelName-retriever, w/o generation task  & 37.9 & 38.6 & 27.8 \\
%     \ModelName, w/o generation task            & 28.7 & 24.7 & 31.4 \\
%     \bottomrule
%     \end{tabular}
% }
%     \caption{Ablation on the pre-training steps used by \ModelName, described in \cref{sec:pretrain}, measured using EM on NQ, TQA, and WQ: we analyse the impact of removing all, auto-encoding, and generation tasks from \ModelName's pre-training phase.}
%     \label{tab:pretrain_results}
% \end{table}

We conduct ablation study on the pre-training steps and the results are shown in \cref{tab:pretrain_results}.
Without fine-tuning, the pre-trained \ModelName outperforms fine-tuned T5-large on NQ and TQA, and has a competitive result on WQ.
When we remove the auto-encoding (KAE and VAE) tasks, the performance on NQ and WQ drops significantly (36.7 $\rightarrow$ 12.9 on WQ). 
Ablating the generation task results in substantially worse EM on NQ and TQA (44.4 $\rightarrow$ 24.7 on TQA)
% When it pre-trains with auto-encoding task, the performances increase in three datasets.
% 
The ablation results demonstrate that both auto-encoding task and generation task are crucial to \ModelName's performance.
Without all the pre-training tasks, \ModelName perform very poorly, and even worse than T5-base baseline.
This may be due to the fact that the key-value memory is not well learned and hence incorporating them will introduce noise to the model, thus leads to poor predictions.

\subsection{Qualitative Analysis} \label{sec:qualitative}
% Examples from NQ and 

\cref{tab:nq_and_wow} shows some examples from NQ and WoW.
The presented QA pairs correspond to the top-5 retrieved dense key-value pairs.
In NQ, we can see that \ModelName retrieves useful key-value and generates correct answer from the first example.
Different from retrieval-only models that only output the top-1 retrieved QA, \ModelName conducted some sort of reranking, and the decoder manages to use the right key-value to generate the answer.
In another example presented in \cref{tab:cases}, it demonstrates that \ModelName's output is not always from retrieved values.
It will ignore the irrelevant key-value pairs, also uses evidences from keys, which are impossible for retrieval-only models. 

In the example from WoW, it requires using the fine-grained knowledge \emph{19th century} to generate response.
We can see that \ModelName retrieves context-related key-value pairs, and mainly uses the two underlined evidences to generate response.
% But it is hard for a basic parametric model to memorise all these fine-grained knowledge and generate such knowledgeable response without grounding on external knowledge.
In contrast, T5-base generates hallucinated response, producing the wrong time ``18th century''.
% \footnote{The earliest recording of music known to exist was in 19th century.}
This shows that, with memory augmentation, \ModelName generates a more faithful and informative response than T5-base.

More examples can be found in \cref{tab:cases} in the appendix. 
We find that \ModelName retrieves useful key-value pairs and makes full use of them to generate answers.
This analysis also demonstrates the interpretability of \ModelName, and the feasibility of only using dense key-value embeddings to provide knowledge.

% \begin{table}
% \renewcommand\arraystretch{1.2} 
% \setlength\tabcolsep{4pt}
% \centering
% \resizebox{\columnwidth}{!}{
%     \begin{tabular}{ll}
%     \toprule
%     \textbf{Q}: who plays the judge in drop dead diva? & \textbf{A}: \emph{Lex Medlin}  \\ 
%     \textbf{Generate}: \emph{Lex Medlin} \\ \midrule
%     \multicolumn{2}{c}{\textbf{Retrieved Key-Values}} \\ \midrule
%     \textbf{Q}: who plays jane on drop dead diva? & \textbf{A}: Brooke Elliott \\
%     \textbf{Q}: who played fred in drop dead diva? & \textbf{A}: Beverly Hills\\
%     \textbf{Q}: who played empress katia on drop dead diva? & \textbf{A}: Tobin \\
%     \textbf{Q}: who plays judge french in drop dead divorce season 4? & \textbf{A}: \underline{\emph{Lex Medlin}} \\
%         \textbf{Q}: who played ian holt on drop dead diva? & \textbf{A}: Jeffrey Pierce \\
%     \bottomrule
%     \end{tabular}
% }
%     \caption{An example from Natural Question test-set. Noting that \ModelName only retrieves and integrates dense key-value pairs and not accesses the text QAs.
%     % , we presents the text of key-value pairs here.
%     }
%     \label{tab:pre-train_results}
% \end{table}

\begin{table}
\renewcommand\arraystretch{1.2} 
\setlength\tabcolsep{4pt}
\centering
\resizebox{\columnwidth}{!}{
    \begin{tabular}{ll}
    \toprule
    \multicolumn{2}{c}{\textbf{Natural Question}} \\ \midrule
    \textbf{Q}: who plays the judge in drop dead diva? & \textbf{A}: \emph{Lex Medlin}  \\ 
    \textbf{\ModelName}: \emph{Lex Medlin} \\ \midrule
    \multicolumn{2}{l}{\textbf{Retrieved Key-Values}} \\
    \textbf{Q}: who plays jane on drop dead diva? & \textbf{A}: Brooke Elliott \\
    \textbf{Q}: who played fred in drop dead diva? & \textbf{A}: Beverly Hills\\
    \textbf{Q}: who played empress katia on drop dead diva? & \textbf{A}: Tobin \\
    \textbf{Q}: who plays judge french in drop dead divorce season 4? & \textbf{A}: \underline{\emph{Lex Medlin}} \\
        \textbf{Q}: who played ian holt on drop dead diva? & \textbf{A}: Jeffrey Pierce \\
    % \bottomrule
    \end{tabular}
}
\resizebox{\columnwidth}{!}{
    \begin{tabular}{ll}
    \toprule
    \multicolumn{2}{c}{\textbf{Wizard-of-Wikipedia}} \\ \midrule
    \multicolumn{2}{l}{\textbf{Dialogue History}} \\ 
    \textbf{Apprentice}: &  I like jazz. \\
    \textbf{Wizard}: & That's great! Jazz ... is originated  in theafrican-american communitie. \\
    \textbf{Apprentice}: & When did it originate? \\ \midrule
    \multicolumn{2}{l}{\textbf{Response} }\\
    \multicolumn{2}{l}{Target: \hspace{0.65em}  Jazz originated in the \emph{late 19th century}} \\
    \multicolumn{2}{l}{T5-base: \hspace{0.05em}  It was first recorded in the late 18th century} \\
    \multicolumn{2}{l}{\ModelName: \hspace{0.65em}  It originated in \emph{late 19th century} in \emph{new orleans}} \\
    \midrule    
    \end{tabular}
}
\resizebox{\columnwidth}{!}{
    \begin{tabular}{ll}
    \multicolumn{2}{l}{\textbf{Retrieved Key-Values}} \\    \textbf{Q}: where did the genre of jazz originate? & \textbf{A}: \underline{\emph{New Orleans}}, US \\
    \textbf{Q}: when did jazz music start in the united states? & \textbf{A}: \underline{1920s}\\
    \textbf{Q}: what type of music was jazz originally? & \textbf{A}: dance music \\
    \textbf{Q}: what genre of music does rock come from? & \textbf{A}: blues \\
    \textbf{Q}: what genre of music is hip hop? & \textbf{A}: rap \\
    \bottomrule
    \end{tabular}
}
    \caption{Examples from NQ and WoW. Noting that \ModelName only retrieves and integrates dense key-value pairs, but not accesses the presented text-based QAs.}
    \label{tab:nq_and_wow}
\end{table}

    %     & question: where did the genre of jazz originate? answer: \underline{\emph{New Orleans}}, United States\\
    %     & question: when did jazz music start in the united states? answer: \underline{1920s} \\
    %     % & question: what type of music was jazz originally? answer: dance music\\
    %     % & question: what genre of music does the american jazz orchestra play? answer: jazz\\
    %     % & question: what type of music is jazz? answer: jazz\\
    %     % & question: where did jazz music originate from? answer: New Orleans, United States\\
    %     & question: what genre of music does rock come from? answer: blues\\
    %     % & question: what genre of music is hip hop? answer: rap\\
    %     % & question: when did gospel music start in america? answer: the early 17th century"\\
    %     % & question: what type of music is popular in america? answer: country\\ \midrule
    % \bottomrule
    % \end{tabular}

\section{Conclusions}

% In this work, we propose an ....
% The model does what
% Experimental results shows that
% In the future, ....

In this work, we propose the Efficient Memory-Augmented Transformer (\ModelName) that combines the strength of parametric model and retrieval-augmented model. 
It encodes external knowledge into a key-value memory and exploits the fast MIPS search for memory querying.
We introduce pre-training tasks to learn better key-value representations and integration of multiple memory slots into transformer. %
Experiments on knowledge intensive tasks, including open-domain question answering, dialogue and long-form question answering, show both the accuracy and speediness of \ModelName.
In the future, we will seek to improve integrate more diverse knowledge into the memory and generalise our method to more downstream tasks.

\clearpage

\section*{Limitations}  % mandatory section

One limitation is that the memory retrieval module requires weak supervision to train with. 
This may mean that we define different weak supervision labels when apply to different downstream tasks.
One could use end-to-end training techniques such as the ones proposed by \citet{HINDSIGHT, RAG}, to train the memory retrieval module with gradients from the decoder, and we leave this as future work.
% First, our training method is not end-to-end, where we design the retrieval objective to train the memory retrieval additionally.
% 
% \citet{HINDSIGHT, RAG} recently propose the end-to-end training methods for retrieval-augment generation model.
%
% We will refer to their works and research to improve the training method in the future work.
%
%
Another potential limitation is that, we need to store the dense key-value memory $\mathbf{M}$, which requires around 300GB CPU RAM.
But since it is relatively easy to get machine with more CPU RAM than GPU memory, and the fact that most deep learning workstations can reach this requirement, we believe this is not too much a constraint.
%
% Thus, the hardware requirement is still acceptable.
%
Besides, we can use LRU cache to save RAM in low memory resource situations.

\section*{Acknowledgements}
Pasquale and Pontus were partially funded by the European Union’s Horizon 2020 research and innovation programme under grant agreement no. 875160, and by an industry grant from Cisco. Baotian Hu and Yu Zhao were funded by grants: Natural Science Foundation of China (No.62006061), Stable Support Program for Higher Education Institutions of Shenzhen (No. GXWD20201230155427003-20200824155011001). The authors would also like to thank Patrick Lewis, Wenhu Chen, and Zetian Sun for their help and feedback.

\bibliography{emnlp2022}
\bibliographystyle{acl_natbib}

\clearpage

\appendix

\section{Data Efficiency}
\begin{figure*}[hb]
\centering
    \begin{minipage}[t]{0.28\linewidth}
        \centering
        \includegraphics[width=\textwidth]{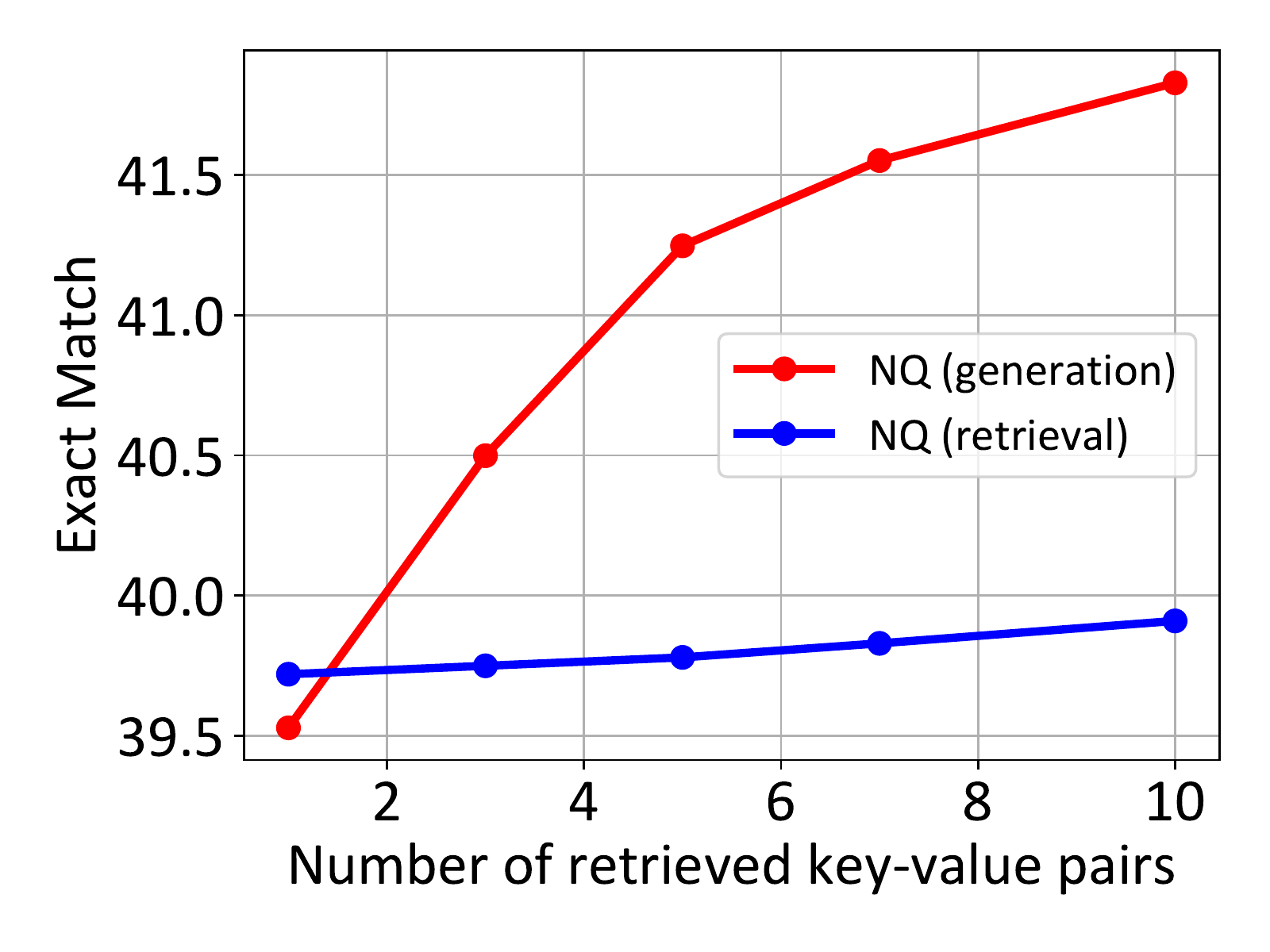}
        \footnotesize \centerline{a) Natural Questions}
    \end{minipage}%
    \begin{minipage}[t]{0.28\linewidth}
        \centering
        \includegraphics[width=\textwidth]{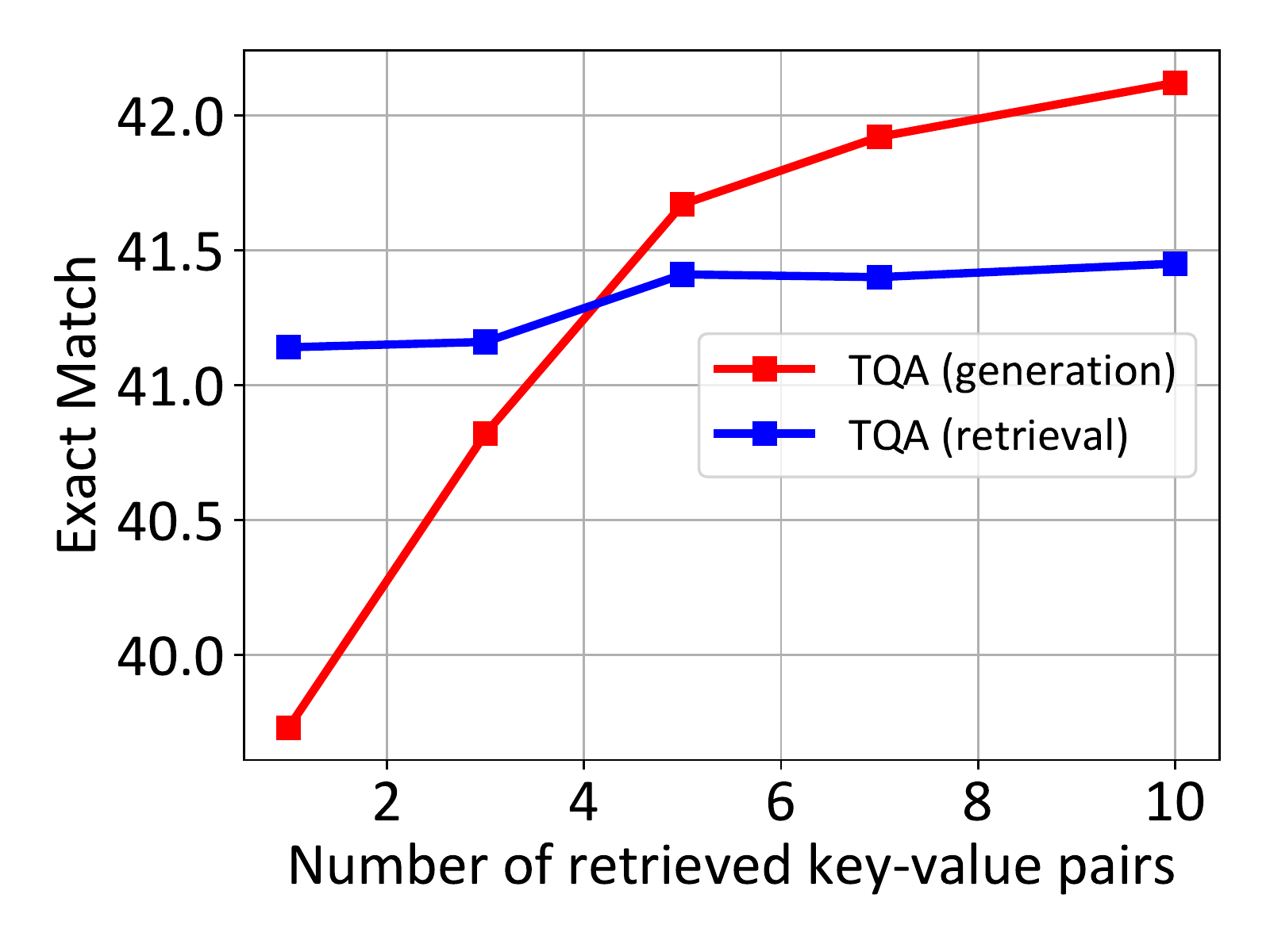}
        \footnotesize \centerline{b) TriviaQA}
    \end{minipage}
    \begin{minipage}[t]{0.28\linewidth}
        \centering
        \includegraphics[width=\textwidth]{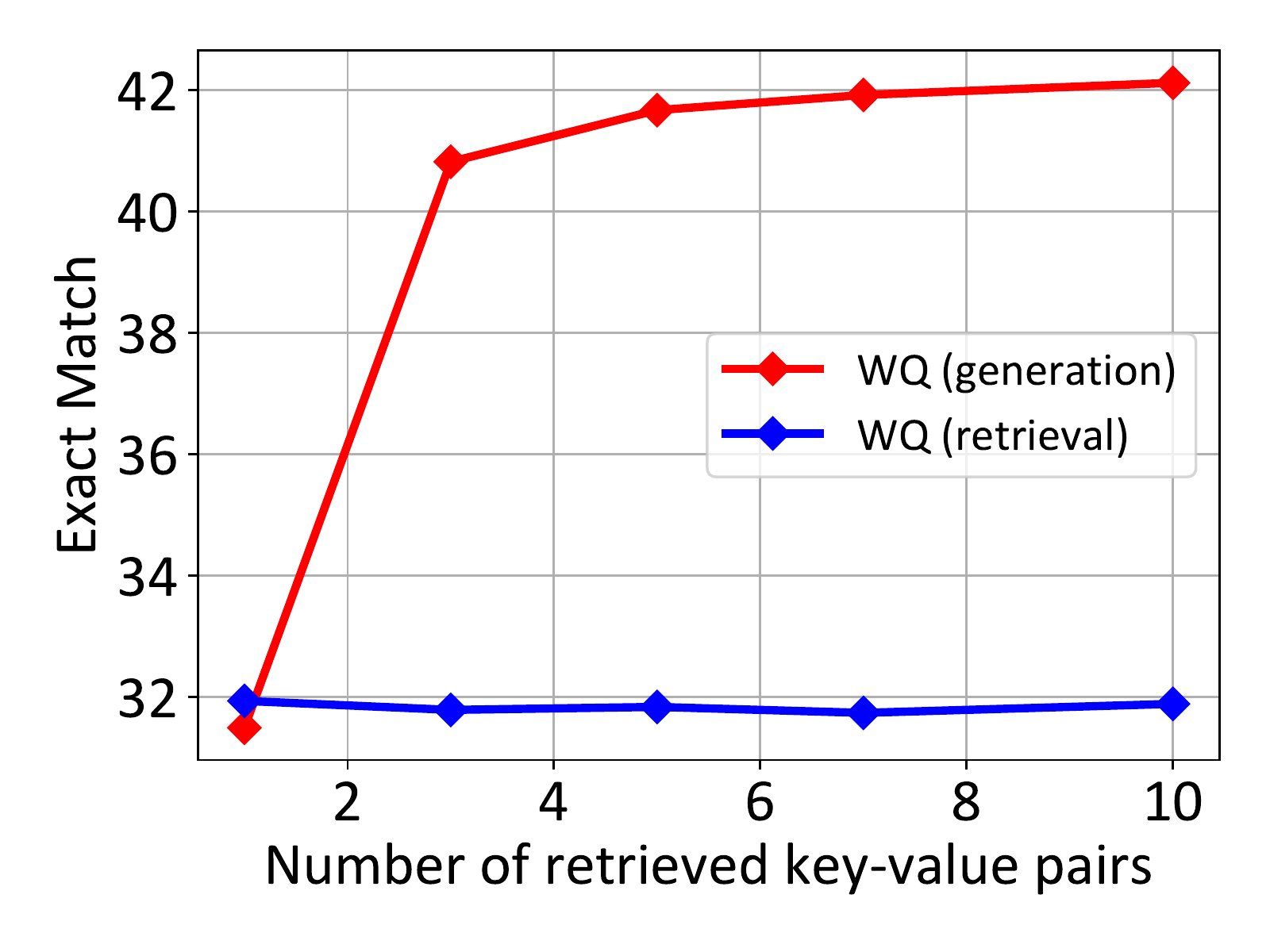}
        \footnotesize \centerline{c) WebQuestions}
    \end{minipage}
    \caption{Analysis of how changing the number of retrieved key-value pairs influences the downstream Exact Match accuracy on several ODQA datasets.} \label{fig:nkvem}
\end{figure*}

\begin{figure}[h]
    \begin{minipage}[t]{\linewidth}
        \centering
        \includegraphics[width=0.68\columnwidth]{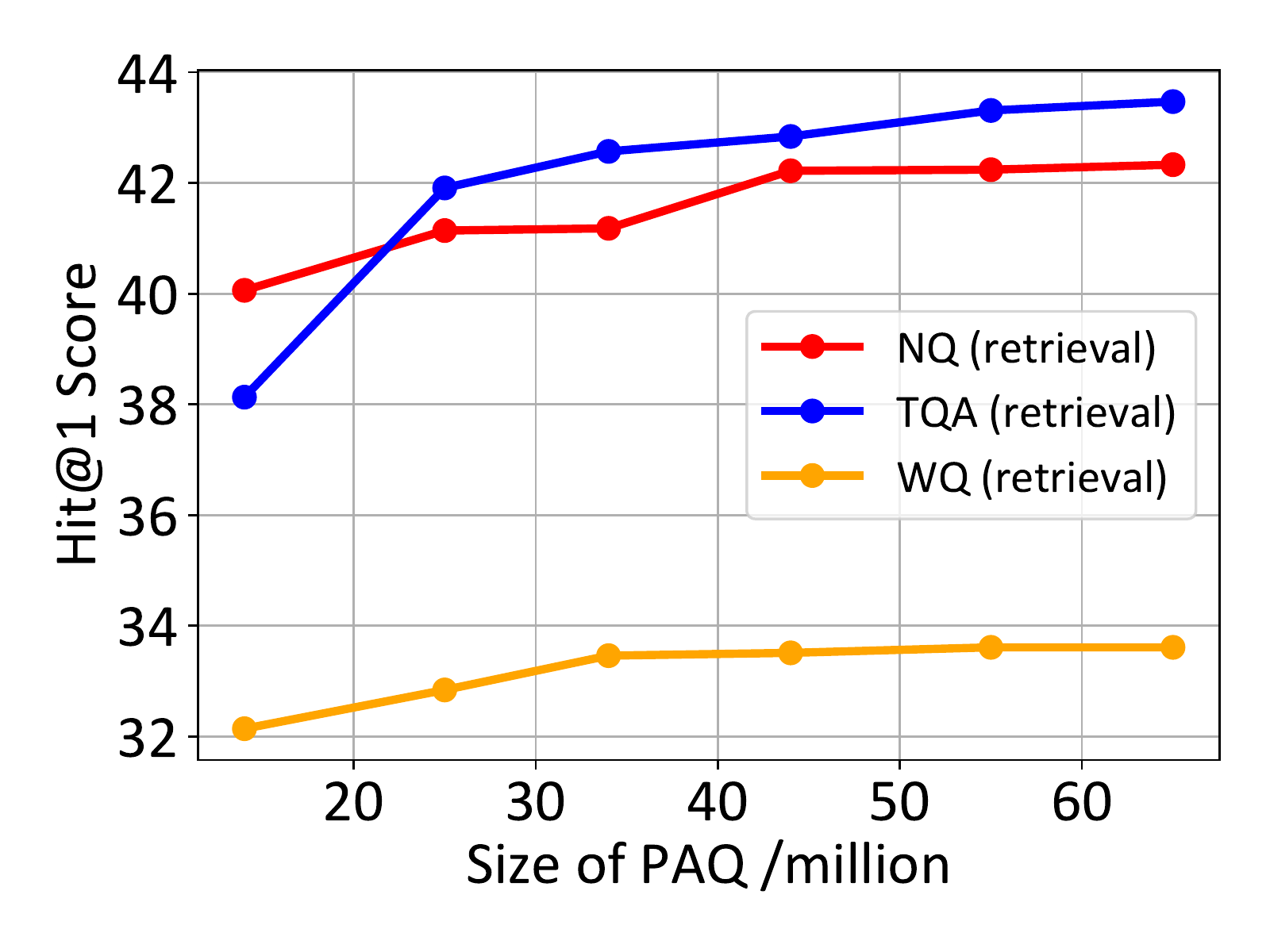}
        \footnotesize \centerline{a) \ModelName-retriever Hit@1}
    \end{minipage}
    \begin{minipage}[t]{\linewidth}
        \centering
        \includegraphics[width=0.68\columnwidth]{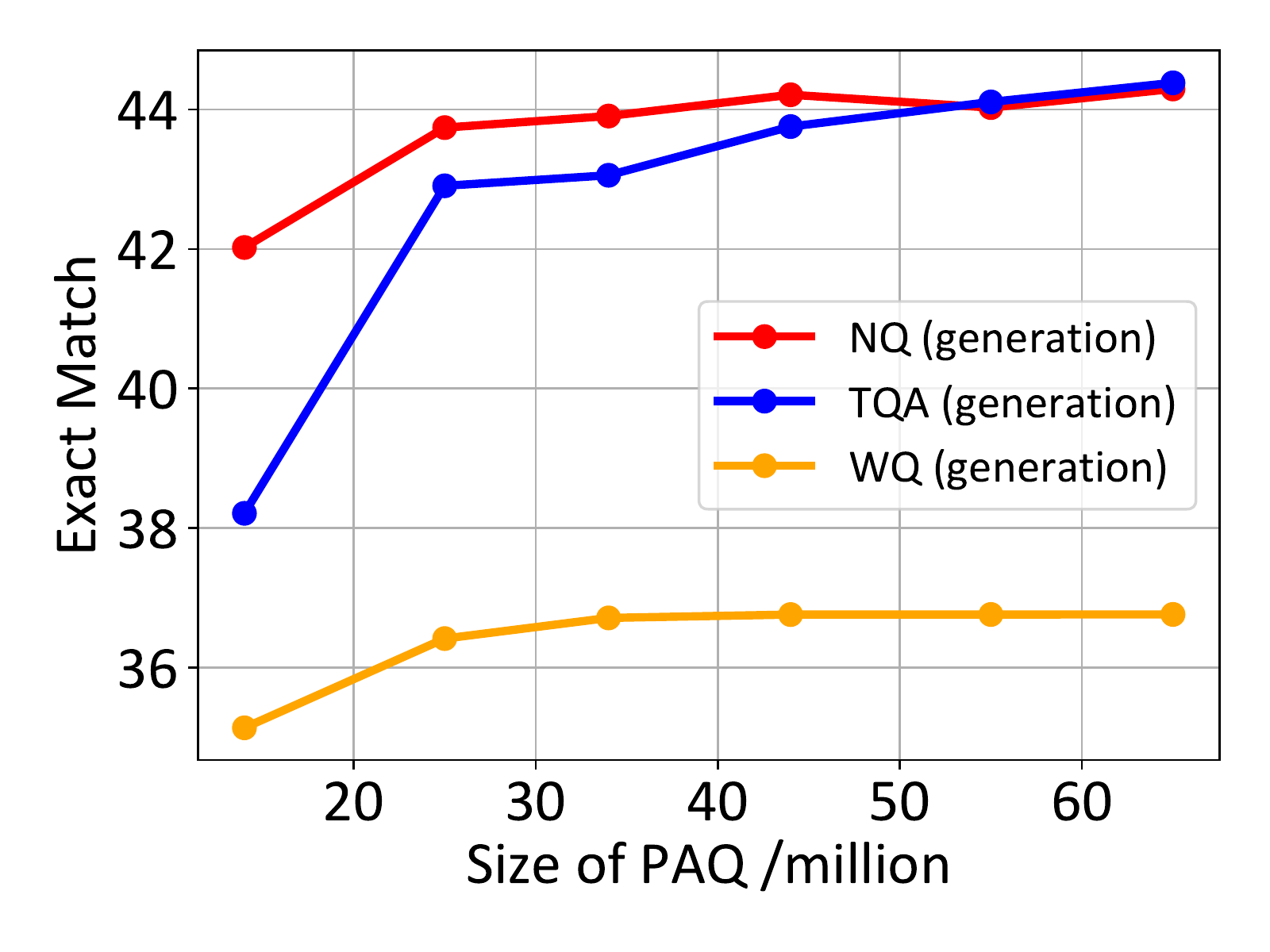}
        \footnotesize \centerline{b) \ModelName-generator EM}
    \end{minipage}
    \caption{Analysis of how the number of PAQ entries used to populate the memory $\mathbf{M}$ influences the downstream predictive accuracy on several ODQA datasets.} \label{fig:size}
\end{figure}

\cref{fig:nkvem} shows how the number of retrieved key-value pairs from PAQ-L1 influences the downstream EM score on Natural Questions, TriviaQA, and WebQuestions.
The results show that, as the number of retrieved memory entries increases, \ModelName's EM score monotonically increases, which indicates that the model can handle noise in the retrieved memory entries and benefit from larger number of retrieved memory entries.
In \cref{fig:size} we analyse the scaling effects induced by using larger subsets of PAQ for creating the key-value memory $\mathbf{M}$.
The results demonstrate that \ModelName's predictive accuracy increases with the number of PAQ questions across all considered ODQA datasets.

\section{Hyperparameters} \label{sec:hyper}
\paragraph{Model Settings}
The length of \prefix is 2 in \ModelName.
\ModelName contains 225M parameters, and T5-base contains 221M parameters.
The memory cache size is set to $384$ in all downstream tasks. The retrieval loss weight and generation loss weight are both set to $1$.
%
%\yw{Zhaoyu please add the hyperparameters and hardware configurations here.}
%
\paragraph{Pre-Training}

We pre-train for $5$ epochs on PAQ-L1, using learning rate warm-ups for the first $5000$ training steps to $10^{-4}$, and linear rate decay in the remaining steps.
For each QA in PAQ-L1, we use RePAQ to retrieve $10$ relevant QAs from PAQ-L1.
To force the model use relevant QAs' information, we sample $10\%$ examples to retain itself in the relevant QA set.
The weights of auto-encoding loss and generation loss is set to $0.5$ and $1.0$.

\paragraph{ODQA}
For NQ and TQA, the learning rate warm-ups for the first $1000$ steps to $5\times10^{-5}$, and linear rate decay in the remaining steps.
For WQ, the learning rate is fixed to $4\times10^{-5}$ during training.
We fine-tune $30$ epochs on ODQA tasks, using early stop with patients of $8$ epochs.
We use greedy decoding algorithm to generate answers.

\paragraph{WoW}
We fine-tune $20$ epochs on WoW with $8\times10^{-5}$ learning rate. The scheduler is same to ODQA. We use greedy decoding algorithm to generate responses.
%We set the learning rate to $5\times10^{-5}$ in ODAQ tasks and $8\times10^{-5}$ in all other tasks.
%
\paragraph{ELI5}
We fine-tune $8$ epochs on ELI5 with $5\times10^{-5}$ learning rate. The scheduler is same to ODQA. We use beam-sample decoding algorithm to generate answers, where beam-size is $5$, top-k is $64$. We force the model do not generate repeat phrases by setting no\_repeat\_n\_gram to $8$.

\paragraph{Hardware}
The machine used to measure the speed is a machine learning workstation with Intel(R) Xeon(R) Platinum 8358 CPU, 512GB of CPU RAM and one 40GB NVIDIA A100 GPU.

% \clearpage
% \section{Generation Cases} 

\begin{table*}[hb]
\renewcommand\arraystretch{1.2} 
\setlength\tabcolsep{4pt}
\small
\centering
\resizebox{\textwidth}{!}{
    \begin{tabular}{ll}
    \toprule
    \multicolumn{2}{c}{\textbf{Natural Questions}} \\ \midrule
    
    \textbf{Question} & who plays the judge in drop dead diva \\
    \textbf{Answer} & [\emph{Lex Medlin}] \\
    % ground truth answers: life imprisonment, life \\
    \textbf{\ModelName Predict}: & \emph{Lex Medlin} \\
    \multirow[t]{3}{*}{\textbf{Retrieved}} 
    & question: who plays jane on drop dead diva? answer: Brooke Elliott \\ 
    & question: who plays judge french in drop dead divorce season 4? answer: \underline{\emph{Lex Medlin}} \\
    & question: who played fred in drop dead diva? answer: Beverly Hills, California \\
 \midrule
    \textbf{Question} & how long did the menendez brothers get in prison for killing their parents \\
    \textbf{Answer} & [\emph{life imprisonment}, \emph{life}] \\
    \textbf{\ModelName Predict}: & \emph{life} \\
    \multirow[t]{3}{*}{\textbf{Retrieved}} 
    & question: when did the menendez brothers kill their parents? answer: 1989 \\ 
    & question: where did the menendez brothers kill their parents? answer: Beverly Hills, California \\
    & question: who sentenced the menendez brothers to \underline{\emph{life in prison}}? answer: Judge Weisberg \\ \midrule
    
    \textbf{Question} &  how long is a whale shark in meters \\
    \textbf{Answer} & [12.65m, estimated at 9.7m, 9.7m] \\
    \textbf{Predict}: & few meters \\
    \multirow[t]{3}{*}{\textbf{Retrieved}} 
    & question: how long does a whale shark live? answer: 70 to 100 years \\ 
    & question: how long does it take a whale shark to mature? answer: around 30 years \\
    & question: how long does it take a blue whale to dive? answer: 10 minutes \\ \midrule
    
    \multicolumn{2}{c}{\textbf{Wizard-of-Wikipedia}} \\ \midrule
    
    \multirow{2}{*}{\textbf{Dialogue history}} 
        & Wizard: Red the color at the end of the visible light spectrum looks good on everyone. \\
        & Apprentice: I am more of a fan of green. That would leave us only one primary color: Blue.  \\ 
    \textbf{Ground Truth} & But the dominant wavelength of red is approximately 625–740. That's impressive! \\
    \textbf{T5 Predict} & I agree. It is the color between green and red. \\
    \textbf{\ModelName Predict} & it is \emph{color between violet and green} on spectrum of visible light \\ 
    \multirow[t]{3}{*}{\textbf{Retrieved}} 
        & question: what is the next color in this series: green, white, red, green, ? answer: Blue\\
        & question: what is the color of \underline{\emph{light between violet and green}}? answer: Blue \\
        & question: what color looks more blue as it brightens? answer: Violet\\ \midrule
    \multirow{3}{*}{\textbf{Dialogue history}} 
        & Apprentice: I like jazz. \\
        & Wizard: That's great! Jazz is a music genre that originated in the african-american communities. \\
        & Apprentice: When did it originate? \\ 
    \textbf{Ground Truth} & Jazz originated in the \emph{late 19th century} \\
    \textbf{T5 Predict} & It was first recorded in the late 18th century \\
    \textbf{\ModelName Predict} & It originated in \emph{late 19th century} in \emph{new orleans} \\ 
    \multirow[t]{3}{*}{\textbf{Retrieved}} 
        & question: where did the genre of jazz originate? answer: \underline{\emph{New Orleans}}, United States\\
        & question: when did jazz music start in the united states? answer: \underline{1920s} \\
        & question: what genre of music does rock come from? answer: blues\\
    \bottomrule
    \end{tabular}
}
    \caption{More examples of \ModelName's prediction on NQ and WoW.}
    \label{tab:cases}
\end{table*}

\end{document}